\documentclass[a4paper, 10pt, conference]{ieeeconf}
\IEEEoverridecommandlockouts
\usepackage{comment} 
\usepackage{amsmath,amssymb,bm}
\usepackage{subcaption,graphicx}
\usepackage[textsize=footnotesize,color=white]{todonotes}
\usepackage{placeins}
\usepackage{overpic}
\usepackage[final]{changes}

\DeclareMathOperator*{\argmax}{arg\,max}

\title{\LARGE \bf Feedback-based Fabric Strip Folding}
\author{Vladim\'{i}r~Petr\'{i}k and Ville~Kyrki,~\IEEEmembership{Senior~Member,~IEEE}%
\thanks{Authors are with Aalto University, Finland {\tt\small\{vladimir.petrik, ville.kyrki\}@aalto.fi}.}%
\thanks{This work was supported by Academy of Finland, decision 317020.}%
}

\usepackage{regexpatch}
\makeatletter
\xpatchcmd{\@todo}{\setkeys{todonotes}{#1}}{\setkeys{todonotes}{inline,#1}}{}{} 
\makeatother

\begin{document}
    \maketitle
    \thispagestyle{empty}
    \pagestyle{empty}

    \begin{abstract}
    Accurate manipulation of a deformable body such as a piece of fabric is difficult because of its many degrees of freedom and unobservable properties affecting its dynamics.
    To alleviate these challenges, we propose the application of feedback-based control to robotic fabric strip folding.
    The feedback is computed from the low dimensional state extracted from a camera image.
    We trained the controller using reinforcement learning in simulation which was calibrated to cover the real fabric strip behaviors.
    The proposed feedback-based folding was experimentally compared to two state-of-the-art folding methods and our method outperformed both of them in terms of accuracy.
\end{abstract}

    \section{INTRODUCTION}
Feed-forward robotic manipulation of an object requires accurate prediction of the object state evolution under the given robot actions.
This prediction is difficult especially for deformable/soft objects because they have a high degree of freedom and the deformation usually depends on many directly unobservable object properties.
If reliable prediction is not available, feedback-based control can be used to increase accuracy.

\begin{figure}[t]
    \centering
    \includegraphics[width=0.99\linewidth]{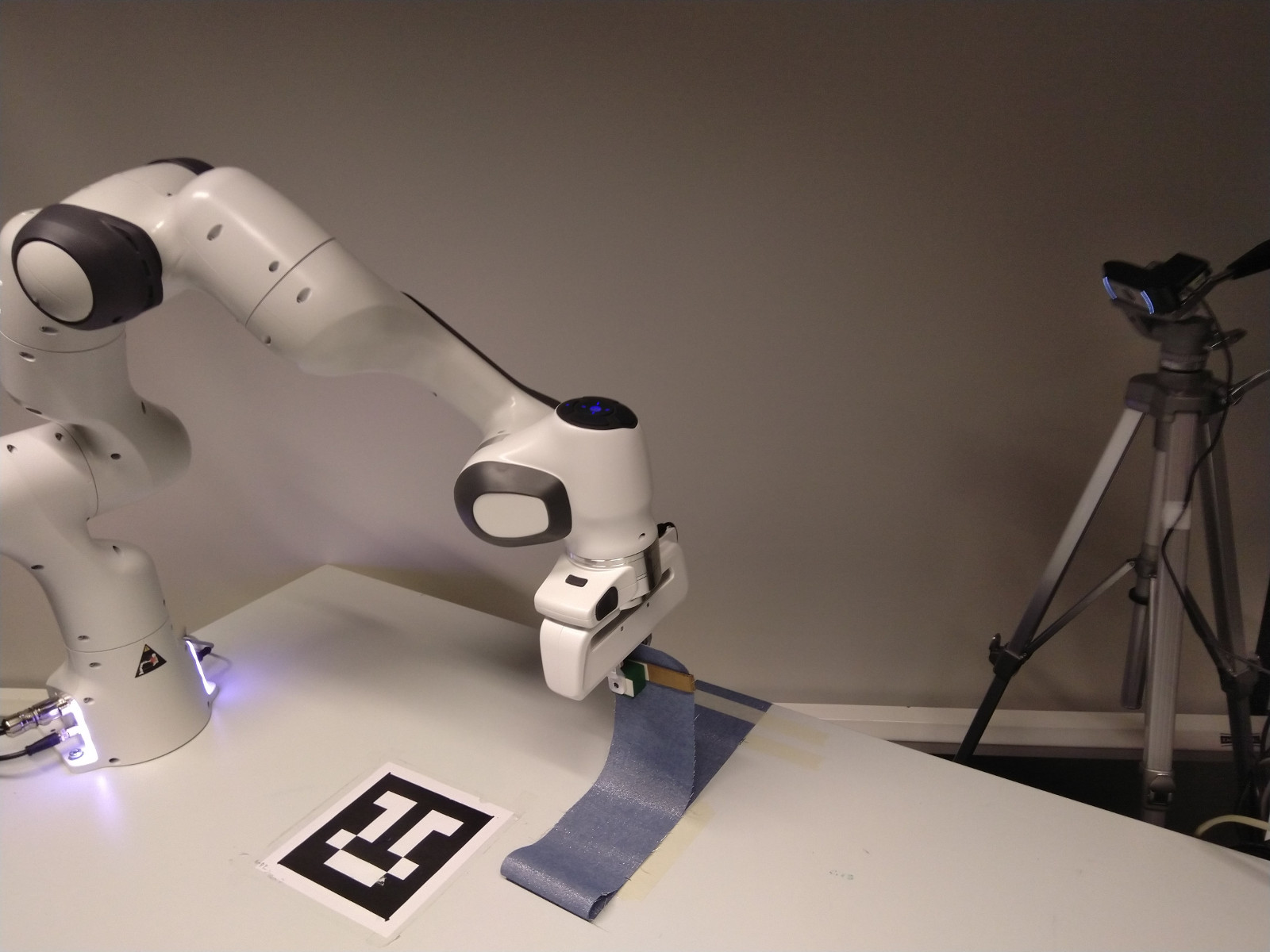}~%
    \caption{
    The \textit{Franka Emika Panda} robot used in the experiment to perform folding of a strip.
    The folding proposed in this paper is based on feedback computed from the images obtained from the calibrated camera.
    }
    \label{fig:intro_img}
\end{figure}

Robotic strip folding visualized in Fig.~\ref{fig:intro_img} is an example of soft material manipulation.
In the folding, a robot grasps one edge of the strip and follows the folding path.
The shape of the folding path affects the accuracy of the folding mainly for two reasons:
(1) the gripper can cause the slipping of the strip on the desk,
(2) the relative position of the upper and lower layers is hard to change after first touch because of friction.

Several approaches for designing folding paths have been proposed in the literature.
Two of them~\cite{BergWAFR2010,PetrikTAROS2015} are based on unrealistic assumptions to design paths which are independent of material properties.
They are easy to compute but ignoring the material properties leads to inaccurate folds.
Other approaches~\cite{LiIROS2015,PetrikIROS2016,PetrikADR2017} take the material properties into account but assume the properties are known in advance or rely on special hardware to estimate the properties in a course of folding and are computationally expensive~\cite{PetrikICRA2018}.

In this paper, we propose to use a feedback-based controller to perform the folding in a closed loop.
The input to the controller is a low dimensional state computed from an image obtained from a calibrated camera.
The controller policy is determined by reinforcement learning (RL) in simulation using domain randomization to account for unobservable properties.
The controller is trained to minimize the misalignment of layers after the touch.
The controller is experimentally evaluated by real robotic folding for various strip materials.

The contributions of the paper are:
(1) a vision feedback-based controller for fabric strip folding,
(2) calibration of a simulation model to match the behavior of real fabric strips,
(3) the controller parameters training in simulation by RL algorithm,
(4) experimental comparison of the controller performance w.r.t.\ the state-of-the-art folding methods.

    \section{STATE-OF-THE-ART}
The robotic folding of garments is usually decomposed into individual folds~\cite{MillerIJRR2011,DoumanoglouTRO2016}.
The accuracy of each fold is crucial for obtaining an accurately folded garment in the end because the errors accumulate with each fold.
We study the accuracy of the single fold when applied to one-dimensional single layer fabric strips.

Triangular paths have been proposed for folding~\cite{BergWAFR2010,MillerIJRR2011,DoumanoglouTRO2016}.
This path was derived based on assumptions of infinitely flexible strip and \added{an} infinite friction between the desk and the garment.
A violation of these assumptions results in an inaccurate fold as was experimentally shown e.g.\ in~\cite{PetrikIROS2016} and theoretically analyzed for continuum model of fabric in~\cite{PetrikTMECH2019}.
Another unrealistic assumption of a rigid body with flexibility only in the folding line was used in~\cite{PetrikTAROS2015}.
This assumption results in a circular path which is complementary to the triangular path.
The triangular and circular paths do not depend on the fabric material and are easy to compute but provide inaccurate alignment in the end.

More complex methods for the folding path design are specific to the fabric material and assume that the material properties have been identified prior to folding.
Several methods have been proposed, differing in the type of used model and in the simulation utilized to find the folding path.
A mass-spring model in \textit{Autodesk Maya} simulation was proposed in~\cite{LiIROS2015}, a mechanical physics-based model derived from the \textit{Euler-Bernoulli} beam theory and boundary value problem solver in~\cite{PetrikIROS2016},
and finite element simulation of a model derived from the \textit{Kirchoff-Love} shell theory in~\cite{PetrikADR2017}.
All these methods obtained superior accuracy compared to the triangular and circular paths but they require the material properties to be known before the folding starts.

The previous folding path generation methods do not use feedback in the folding process.
Recent work~\cite{PetrikICRA2018} uses a laser range finder to measure the shape of the strip while folding.
The parameters of a finite-element model~\cite{PetrikADR2017} are then perturbed until the model shape matches the measurement.
The shape fitting is performed after each robot motion to avoid the strip slipping on the desk and to improve the estimation while lifting the strip upward, which requires solving the finite element simulation many times taking several minutes for a single fit.
After the strip is in the upper position the last estimate is used to design the rest of the folding path which is then followed in a feed-forward manner.
The method proposed in this paper differs from~\cite{PetrikICRA2018} by:
(1) we use a standard camera instead of specialized accurate laser range finder which simplifies the calibration process,
(2) we use the feedback during the whole motion,
(3) our method produces a controller which computes the action for the robot quickly and can be used in real time.

Recently, several learning-based approaches to robotic folding were proposed,
e.g.~\cite{jia2018learning,matas2018sim,sannapaneni2017learning,colome2018dimensionality,yang2017repeatable}.
However, these methods have been tested only on a single fabric material and the accuracy has not been measured.

    \section{FABRIC STRIP MODEL}
To learn a folding policy in simulation we require models of strips with realistic dynamics.
In this section, we first describe the phenomena of real fabric strips which affect the folding accuracy.
We then propose a model for simulating these phenomena.
The treatment is restricted to single layer strips.

\subsection{Real Fabric Strip Behavior}
\begin{figure}[t]
    \centering
    \begin{overpic}[width=.45\textwidth,tics=10]{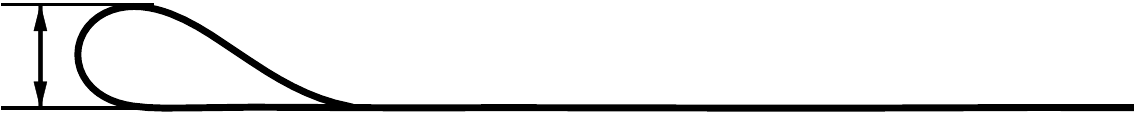}
        \put(0, 4){$h$}
    \end{overpic}
    \caption{
    The folded shape of the strip.
    The maximum height of the folded strip~$h$ depends on the fabric material.
    }
    \label{fig:folded_shape}
\end{figure}

\replaced{A s}{S}hape of a folded fabric strip is visualized in Fig.~\ref{fig:folded_shape}.
The maximum height of the folded strip (denoted~$h$ hereinafter) depends on the fabric material.
Height~$h$ was used in~\cite{PetrikIROS2016} to estimate \textit{weight-to-stiffness-ratio} for continuum model of the strip.
It was then used to compute the folding path for the particular strip material.
Therefore, the folding path depends on~$h$.

The accuracy of \deleted{folding a} strip \added{folding} is affected mainly by the strip slipping and by the position of the first touch of the strip with itself.
The slipping occurs if the horizontal force created by the gripper is larger than the frictional force.
Therefore, it can be avoided by minimizing horizontal force as shown in~\cite{PetrikIROS2016}.
Alternatively, the folding path can be adjusted after the slipping occurs if the slipping is measured online.

The first touch of fabric with itself is shown in Fig.~\ref{fig:state_evolution}.
The position of the first touch measures the alignment of individual layers.
The friction coefficient between the parts of the fabric is usually higher than the friction coefficient between the desk and fabric.
Therefore, after the layers are in contact it is difficult to change the position of the touch without lifting the strip upward and breaking contact.

\begin{figure}[!b]
    \centering
    \includegraphics[width=1.0\linewidth]{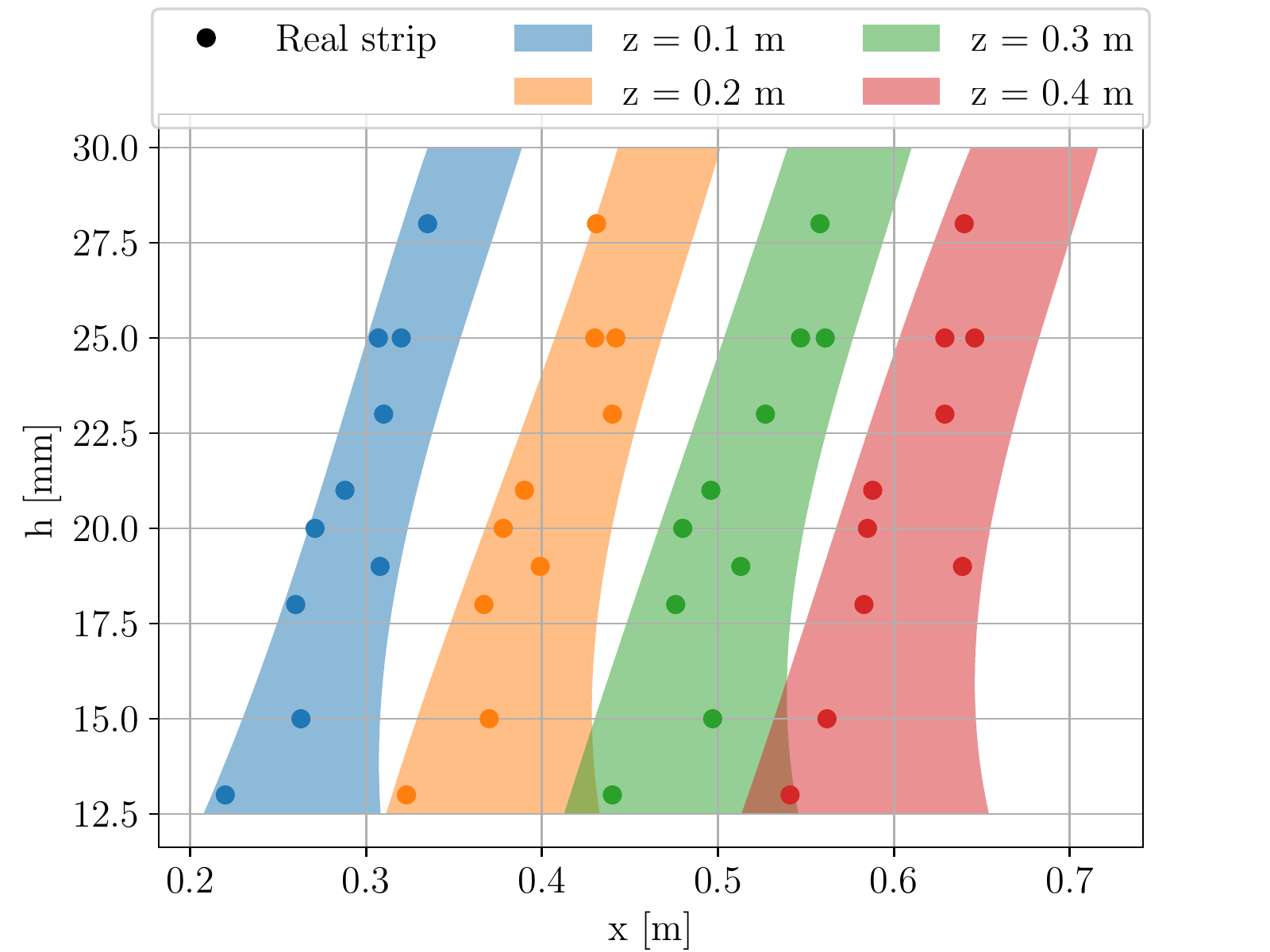}~%
    \caption{
    The position of the gripper~$x$ for various gripper heights~$z$ at which the first touch of fabric layers occurred.
    In the experiment, the $x$-position was changing continuously until the touch occurred.
    Individual colors correspond to different gripper heights.
    The shaded areas correspond to values obtained from \added{the} MuJoCo simulation by continuously changing strip properties.
    The maximum height of the folded strip is denoted by~$h$.
    }
    \label{fig:critical_points}
\end{figure}

Manipulating a strip slowly results in a statically stable equilibrium of the strip in most of the work space.
However, there is a static instability immediately before the touch occurs as documented in~\cite{PetrikTMECH2019}.
This instability results from a dynamic snap-through behavior known from mechanics.
During the dynamic motion, there is another property of fabric material which affects the position of the first touch and therefore the accuracy.
This property was not modeled in the continuum models used for the folding path generation in~\cite{PetrikIROS2016}.
The effect of this property on the first touch position for various gripper paths was measured experimentally in~\cite{PetrikTMECH2019}.
We plot these data in Fig.~\ref{fig:critical_points}.

\subsection{MuJoCo Simulation}
We use \added{a} MuJoCo~\cite{todorovIROS2012mujoco} environment to simulate fabric strips.
The strip is modeled as a one-dimensional sequence of equally spaced and weighted spheres connected by rotational joints.
The joints allow motion of the strip in \textit{xz}-plane only.

We used two parameters of joints which affect the behavior of simulated strip: stiffness and damping.
Other parameters were fixed to constant values and the number of the \replaced{spheres}{elements} was fixed.
We assume homogeneity of the fabric material and use the same values for all joints.
Continuous change of the stiffness and damping parameters can cover both phenomena:
different folded state height as well as different first touch position after passing instability.
The coverage of the phenomena is shown in Fig.~\ref{fig:critical_points}.
It can be seen, that our model is able to cover all real measurements.
The simulation parameters required to replicate these results are shown in Appendix~\ref{app:technical_details}.

\begin{figure}[b]
    \centering
    \includegraphics[width=1.0\linewidth]{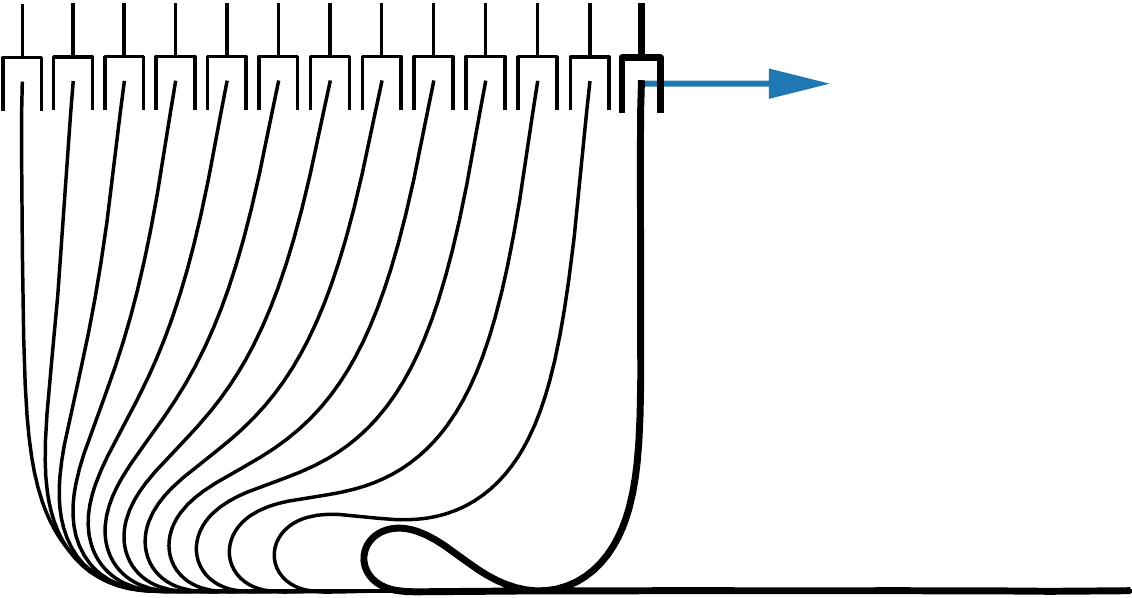}~%
    \caption{
    An evolution of strip states caused by the gripper horizontal motion as visualised by the blue arrow.
    The bold state represents the first state at which the touch of the layers occurs.
    }
    \label{fig:state_evolution}
\end{figure}

Slipping is not measured in simulation directly.
Instead, we fix the non-grasped edge of the strip and measure the force required to hold the edge fixed.
This force is then minimized to avoid slipping similarly to~\cite{PetrikIROS2016}.
This simplifies the simulation because in addition to gripper force the amount of slipping depends on friction coefficient.
Therefore, various friction models would need to be simulated to get the slippage for various real-life conditions.
The example of the model states for one set of strip parameters is shown in Fig.~\ref{fig:state_evolution}.
It should be noted that the force measurement is only used to learn the control policy in simulation and thus not needed in the physical system.

The current structure of the strip model does not simulate the extension of the strip,
i.e.\ the strip length is constant independent of the stretching force.
This is in correspondence with the fact that strips extend negligibly under their own weight as reported in~\cite{PetrikICRA2018}.
However, if an extension is required then the model could be enriched by prismatic joints located between the spheres.
The stiffness of the joints would control the extensibility of the strip.

    \section{Feedback-based Controller}
\begin{figure}[t]
    \centering
    \begin{overpic}[width=\linewidth,tics=10]{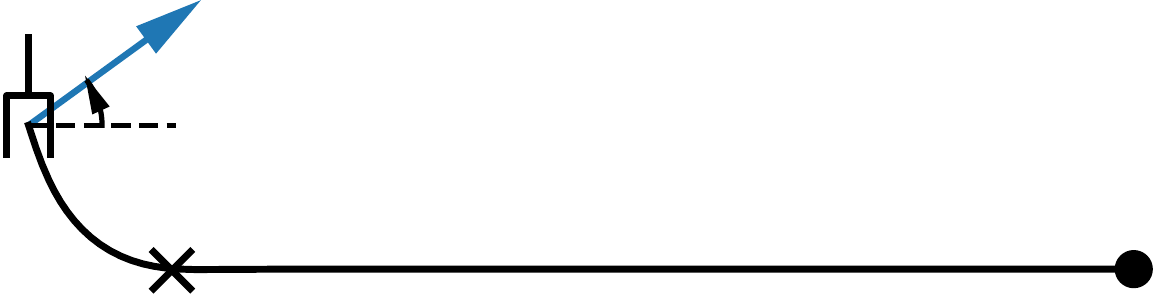}
        \put(11, 17){$\varphi$}
    \end{overpic}
    \caption{
    The visualisation of the state which is used to compute \added{the} action~$ \varphi $ shown by blue arrow.
    The state consists of $x$ and $z$ position of the grasped edge of the strip (origin of the blue arrow) and
    of $x$ position of the first touch of the strip with the desk (cross symbol).
    The \replaced{black dot}{circle} visualises the fixed point which is used to prevent slipping in simulation.
    }
    \label{fig:action_state}
\end{figure}

To control the folding we use a feedback-based controller in a form:
\begin{align}
    \varphi = \pi_{\bm w}(\bm x) \, ,
\end{align}
where $\varphi$ is a one-dimensional action which represents angle between the $x$-axis and relative gripper motion,
\added{and} $\bm x$ is a three-dimensional state containing the gripper position and $x$-coordinate of the first contact of fabric with the desk.
The state and action are visualised in Fig.~\ref{fig:action_state}.
The policy $\pi_{\bm w}$ computes the action~$\varphi$ based on the state~$\bm x$.
The policy is a function approximator parametrized by vector~$\bm w$.
In our setup, we used a neural network with one hidden layer and 20~hidden neurons to represent the policy.
The network uses hyperbolic tangent as an activation function.
The weights of the network parametrize the policy.
First, we will describe how to find parameters~$\bm w$ for a strip with fixed fabric dynamics parameters.
Then, we extend the formulation to generalize the policy across varying parameter values.

\subsection{Single Strip Folding}
The parameters~$\bm w$ are trained in simulation using reinforcement learning.
Thus, we maximize total rewards with respect to policy parameters,
\begin{align}
    \bm w = \argmax\limits_{\bm w} R(\bm \theta) \, , \label{eq:maximization}
\end{align}
where~$ \bm \theta $ represents the strip properties (constant in a single strip folding)
and~$ R(\cdot) $ is the total reward computed from the strip states obtained from the simulation by applying the policy~$ \pi_{\bm w} (\cdot)$.
The total reward can be decomposed into the intermediate reward for $i$-th step~$r_i$ and terminal reward~$ r_f $:
\begin{align}
    R(\bm \theta) = \sum\limits_{i = 0}^{N} r_i(\bm \theta) + r_f(\bm \theta) \, ,
\end{align}
where~$ N $ is the number of simulation steps.
Both the intermediate as well as terminal reward depend on the strip properties~$ \bm \theta $.

\begin{figure}[b]
    \centering
    \begin{overpic}[width=0.8\linewidth,tics=10]{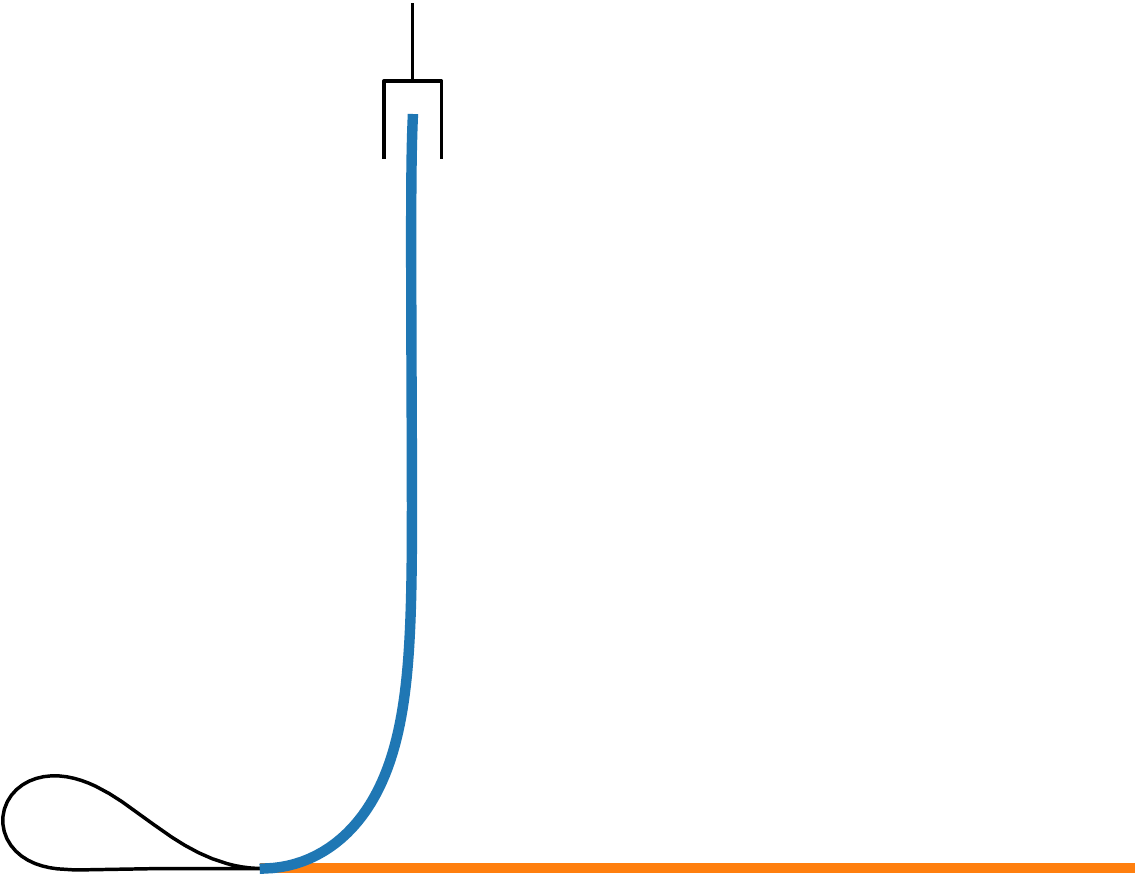}
        \put(31, 35){$l_1$}
        \put(60, 3){$l_2$}
    \end{overpic}
    \caption{
    The displacement of the layers is measured after the strip touches itself.
    The symbol~$ l_1 $ denotes the length of the hanging part of the strip (shown in blue).
    The length of the laying part of the strip measured between the point of layers contact and the fixed end of the strip is denoted by symbol~$ l_2 $ and shown in orange.
    The oriented displacement is computed by~$ d = l_1 - l_2 $.
    }
    \label{fig:layers_misalignment}
\end{figure}

In our scenario, the simulation stops after the touch of layers occurs or after a maximum number of steps~$ H $.
Note, that~$ N $ equals to~$ H $ only in the case when the layers did not touch.
The terminal reward is then computed as
\begin{align}
    r_f =
    \begin{cases}
        -| d |, & \text{if touch occurred,} \\
        -\infty, & \text{otherwise,}
    \end{cases}
    \label{eq:r_f}
\end{align}
where~$ d $ is the misalignment of the layers as visualised in Fig.~\ref{fig:layers_misalignment}.
The intermediate rewards represent the force~$ f_x $  required to hold the strip in order to prevent slipping:
\begin{align}
    r_i = - a \, \frac{|f_x|}{N} \, ,
\end{align}
where~$ a $ is a scale factor manually selected to give higher importance to the terminal reward.

The infinite value of terminal reward in Eq.~\eqref{eq:r_f} can be used if policy initialization ends in the state where layers touch.
Otherwise, reward function gradient does not provide information for the optimization of Eq.~\eqref{eq:maximization}.
In our implementation, we used reward shaping to guide the optimization towards the touch \added{by enforcing horizontal gripper motion as visualised in Fig.~\ref{fig:state_evolution}}.
\deleted{Instead of infinite value we used the negative distance between the gripper position and a fixed point located behind the holding edge.}

\begin{figure}[t]
    \centering
    \includegraphics[width=\linewidth]{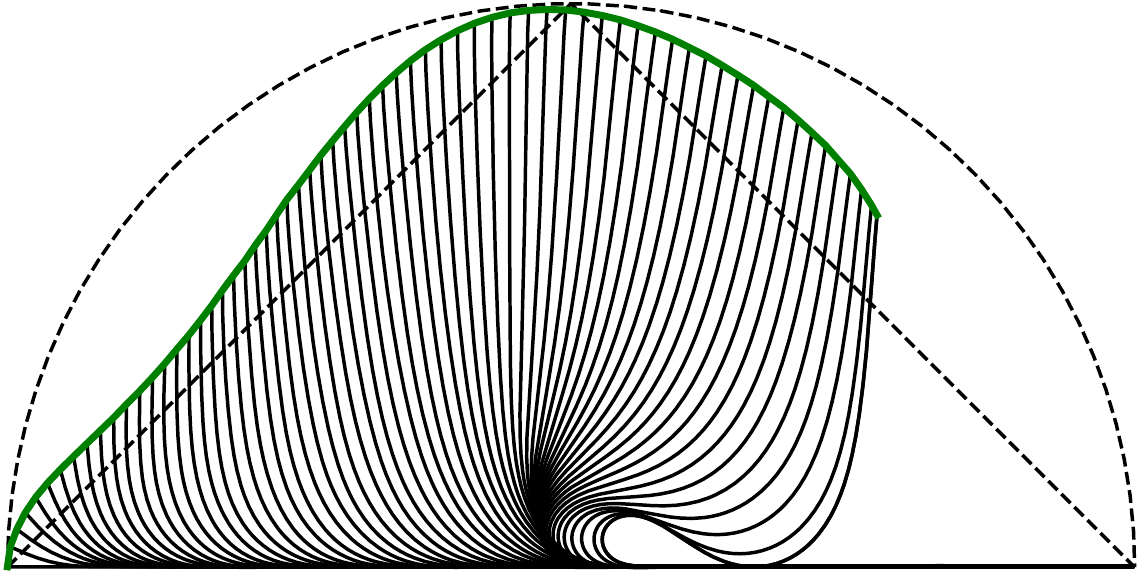}~%
    \caption{
    Evolution of strip state (solid lines) obtained by following the learned control policy.
    The green curve shows the followed path.
    The triangular and circular paths (dashed lines) are shown as a reference.
    }
    \label{fig:policy_single_strip}
\end{figure}

For solving the optimization in Eq.~\eqref{eq:maximization}, we used a modified version of BlackDROPS~\cite{blackdrops} where we removed the Gaussian process model fitting and optimized the reward function in simulation directly with CMA-ES~\cite{cmaes}.
The optimized policy is visualized in Fig.~\ref{fig:policy_single_strip}.

\subsection{Domain Randomization}
\begin{figure}[b]
    \centering
    \includegraphics[width=.8\linewidth]{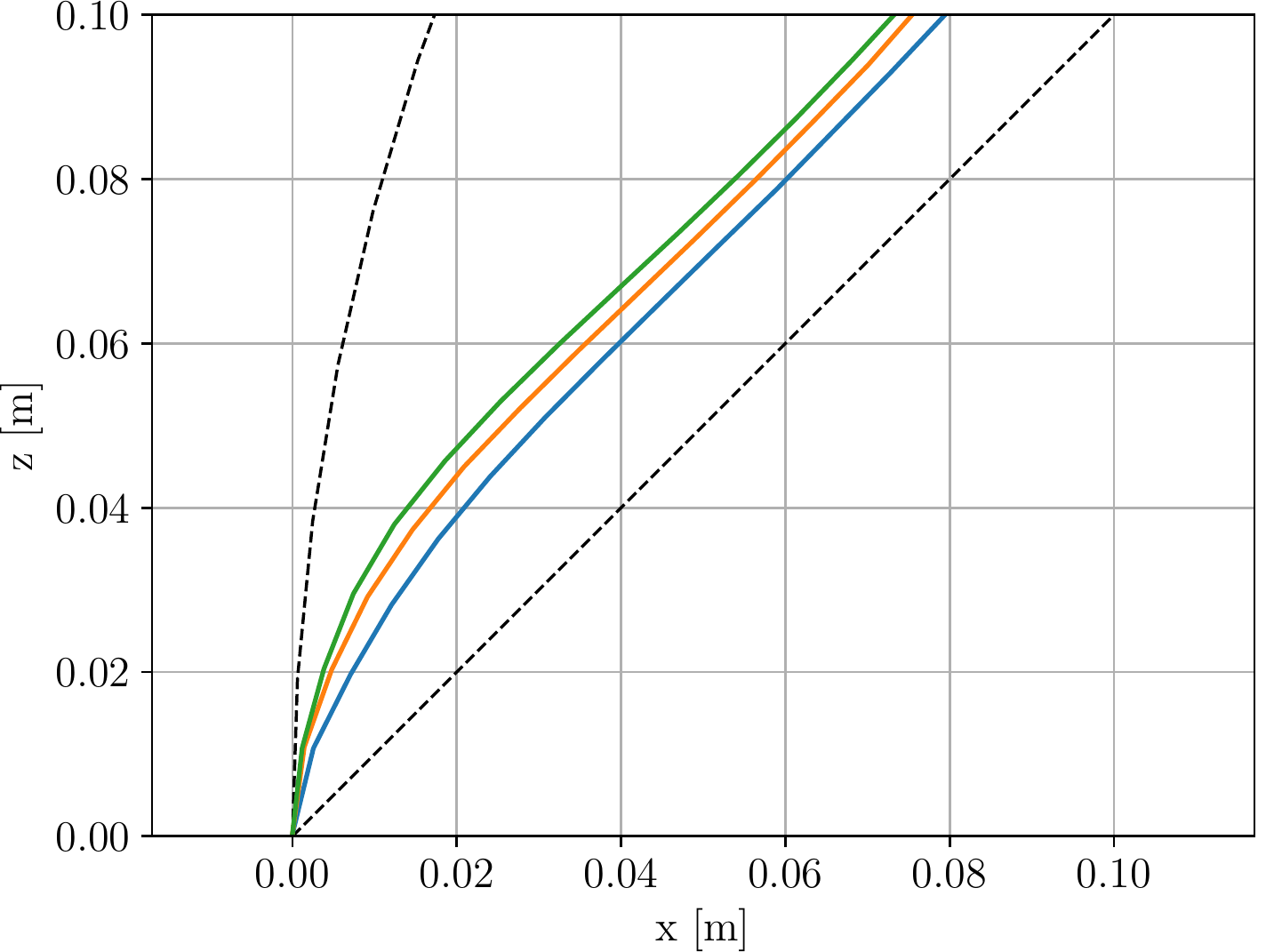}~%
    \caption{
    The policy adapts the folding path based on the feedback in order to minimize the horizontal force.
    Different colors represent paths obtained for different values of stiffness s.t. blue was obtained for the lowest stiffness and the green for the highest stiffness value.
    }
    \label{fig:policy_adaptation}
\end{figure}

The previous approach requires the strip properties~$ \bm \theta $ to be known or directly observable.
However, our goal is to find a policy which is independent on the strip properties.
One possibility of finding such a policy is to use domain randomization~\cite{peng2017sim,chen2018deep}.
Domain randomization for our task may be achieved by modifying the optimization~\eqref{eq:maximization} into:
\begin{align}
    \bm w = \argmax\limits_{\bm w} E_{\bm \theta \sim p(\bm \theta)} \left[  R(\bm \theta) \right] \, , \label{eq:maximizationDR}
\end{align}
where~$ E $ stands for the expectation computed over the parameters given by the prior distribution~$ p(\bm \theta) $.
The prior distribution encodes the strip properties used to generate shaded areas in Fig.~\ref{fig:critical_points}.
By solving Eq.~\eqref{eq:maximizationDR} we obtain the policy which is independent of the strip properties.
The resulting policy adapts to different strips as visualized in Fig.~\ref{fig:policy_adaptation}.

    \section{EXPERIMENTS}
\subsection{\added{Testbed Description}}
\begin{figure}[b]
    \centering
    \includegraphics[width=1.0\linewidth]{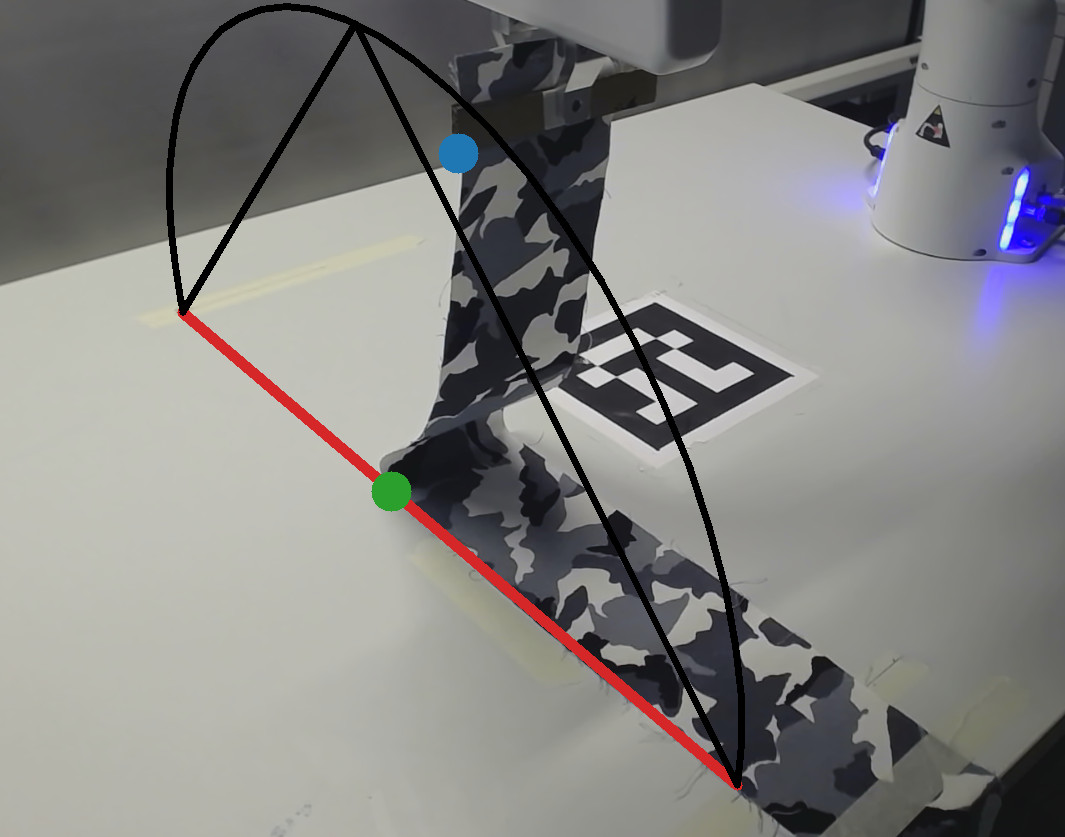}~%
    \caption{
    Image of the fabric used for the feedback control.
    The red line is used for line search of the first touch between the strip and the desk.
    The first touch is shown by the green dot.
    The blue dot shows the projected gripper position.
    The black curves are projected shapes of triangular/circular paths.
    }
    \label{fig:camera_image}
\end{figure}

To perform real experiments we used the \textit{Franka Emika Panda} robot shown in Fig.~\ref{fig:intro_img} and a camera to measure the state.
We estimated a homography~\cite{hartley2003multiple} between the undistorted image plane and the $xz$-plane aligned with the edge of the strip.
The homography was used to project detected position of the first touch between the strip and the desk into the robot coordinate system.
An example of image taken from the camera is shown in Fig.~\ref{fig:camera_image}.

\added{
For the detection of the point at which the strip touches the desk, we performed a line search on the line which corresponds to the strip in the unfolded state.
The colors on the line were compared to the desk color to detect the touch position.
Because of the camera position, it is guaranteed that no other part of the strip will occlude the line in image space before the point of contact.
}

\deleted{For the detection of the point at which the strip touches the desk, we performed a line search and compared the colors on the line with the color of the desk.}
\deleted{The line used for the line search corresponded to the strip in the unfolded state and was computed by the estimated homography.}
\deleted{The searching starts from the grasped edge position.
If the color of the pixel at the line differs from the desk color more than a manually specified threshold, then we consider that point to be the first contact of the strip with the desk.
Because of the position of the camera, it is guaranteed that no other part of the strip will occlude the line in image space before the point of contact.
}

\subsection{\added{Strip Folding}}
\begin{figure}[t]
    \centering
    \includegraphics[width=1.0\linewidth]{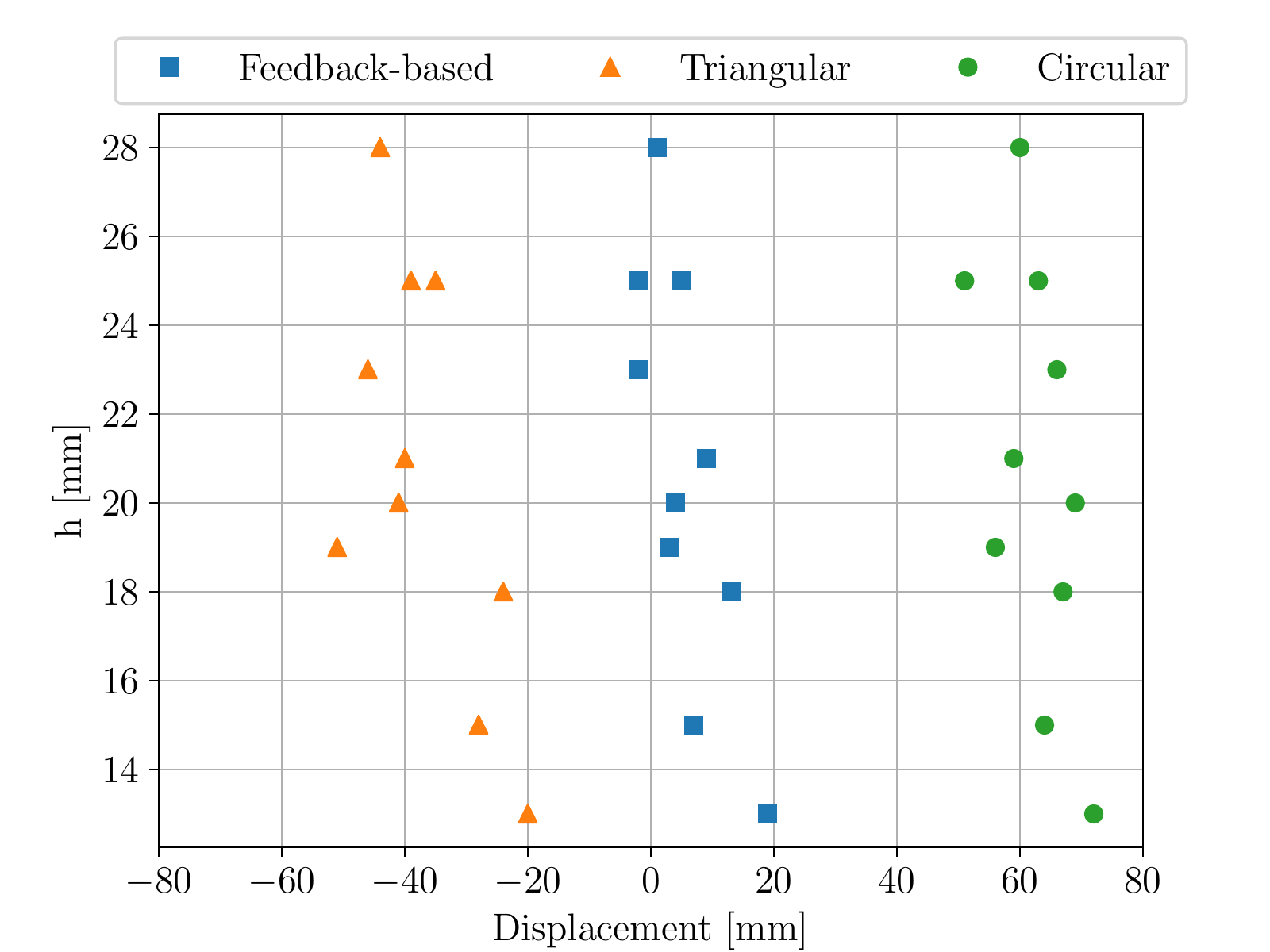}~%
    \caption{
    The measured displacement of the layers after the touch occur\added{ed}.
    The zero displacement corresponds to aligned layers which is the goal of folding.
    }
    \label{fig:displacement}
\end{figure}

We measured the misalignment of layers after passing through the instability~\cite{PetrikTMECH2019}.
The non-grasped edge of the strip was fixed to discard the contribution of slipping to the measurement.
The strip length measured between the grasping point and the fixed point was 0.6~m.
We started folding from the 10~cm height and gripper orientation was fixed s.t.\ the gripper pointed downward (Fig.~\ref{fig:camera_image}).
The folding was stopped manually after the touch of layers occurred (i.e.\ after passing through the instability).
In addition to the proposed feedback-based folding, we measured the displacement for triangular and circular paths for ten different fabric materials.
The measured displacements are shown in Fig.~\ref{fig:displacement}.

The measured oriented displacement for the triangular path is negative for all materials with average displacement of $-36.8$~mm, such that during the folding the strip touched itself sooner than desired.
This is in alignment with predictions made in~\cite{PetrikTMECH2019}.
If the folding continues after the touch occurs, the upper layer can slide on the bottom layer or the whole strip will slip on the desk depending on the friction coefficients.
The displacement for the circular path is positive (average displacement $62.7$~mm) for all materials which is also in alignment with theoretical predictions~\cite{PetrikTMECH2019}.

The feedback-based controller outperformed the triangular and circular paths.
The measured displacement was under 20~mm for all materials without knowing the folded strip properties in advance.
This is inferior to performance of methods that assume the strip properties are known.
For the same fabric materials, the performance obtained in~\cite{PetrikMESAS2016} was under 10~mm.
On the other hand, our method does not require manual estimation of the properties as it adjusts the path automatically based on the measured feedback.

The displacements in Fig.~\ref{fig:displacement} were measured once per strip.
We tested the repeatability of the folding process for one selected strip which we folded 5~times per folding method.
The differences between individual measurements were maximally 1~mm for all tested methods: the feedback-based, triangular, and circular.
The measured differences are lower than the expected accuracy of the whole testbed.
Therefore, we decided to perform one experiment per strip only.

\subsection{\added{Simulated Displacement}}
\begin{figure}[t]
    \centering
    \includegraphics[width=1.0\linewidth]{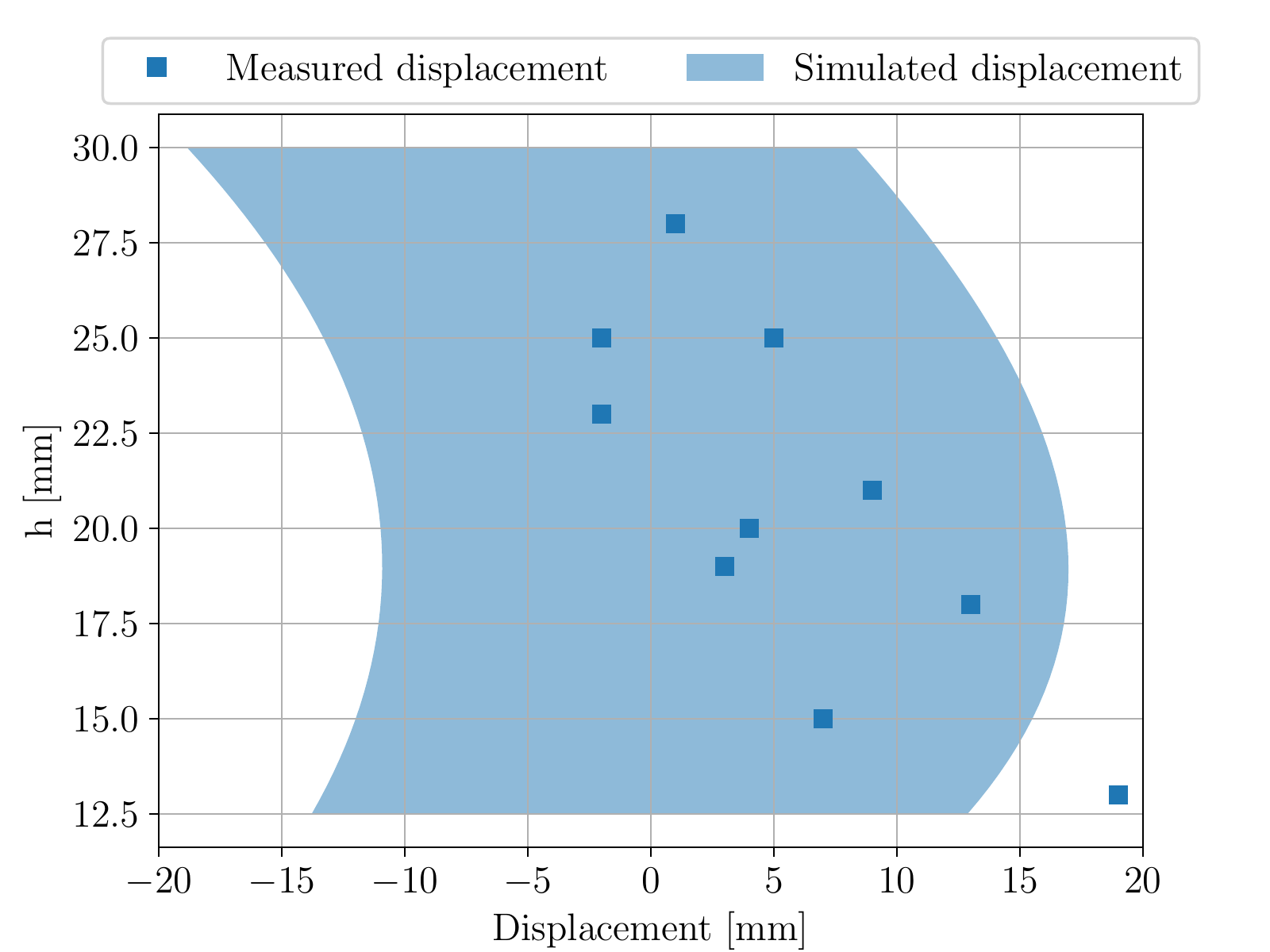}~%
    \caption{
    The comparison of measured displacement (squares) for fabric strips with the simulated displacement for the whole range of simulated strip properties (shaded area).
    }
    \label{fig:sim_displacement}
\end{figure}

To analyze why the feedback-based folding does not perform as well as methods tuned to a particular fabric, we performed the folding in simulation for various strip parameters and measured the displacement.
The obtained results compared to real measurements are shown in Fig.~\ref{fig:sim_displacement}.
It can be seen that zero displacements were not achieved for all strips in the simulation.
This can be caused by the policy representation not being sufficiently complex or by the measured strip state not being sufficiently informative.
Analysis of this issue is the subject of our future work.
All except one real measurements lie in the displacement obtained in simulation.
That indicates that the range of simulated properties needs to be expanded during the training to cover the behavior of that strip.

    \section{CONCLUSIONS}
We proposed a feedback-based approach for robotic strip folding.
In our approach, the low dimensional feedback is measured in an image obtained from a calibrated camera.
We optimized a controller in \added{the} calibrated simulation using reinforcement learning.
The displacement of strip layers and the horizontal force was used in a reward function.
We experimentally verified the controller in real robotic folding for 10~different fabric strips.
Our feedback-based approach outperformed existing approaches that are agnostic to fabric type.

In comparison to the state-of-the-art method for strip property estimation~\cite{PetrikICRA2018},
our approach is based on easily available monocular camera instead of on an accurate laser range finder.
Using the standard camera simplifies the calibration of the testbed and we showed that it can be used to improve the accuracy of folding.
Moreover, the controller is capable of online computation while the state-of-the-art method requires solving an optimization problem in the finite element simulation which takes several minutes.

The measured absolute displacements for the feedback-based folding were under 20~mm for all tested strips.
This performance was obtained without transfer learning by training in simulation using domain randomization technique.
In our future work, we will study how the transfer learning and different policy/state representations could be used to reduce the displacement.

    \begin{appendices}

    \section{}\label{app:technical_details}
    The following parameters were used in MuJoCo simulation~\cite{todorovIROS2012mujoco} to simulate strips:
    distance between spheres: 2~mm,
    spheres diameter: 1~mm,
    spheres mass: 3.3~g,
    strip length: 0.6~m,
    simulation step: 5~ms.
    We used Newton solver with maximum number of 100~iterations.
    The contacts computation between spheres were dissabled.
    The joints stiffness value~$ k $ lies in range: $k \in \left<0.02, 0.3\right>$.
    The minimum damping is stiffness dependent and computed by:
    $b_{\text{min}} = 3 \cdot 10^{-3} k + 3.5 \cdot 10^{-4} $.
    The maximum damping is 50-times larger than minimum damping for each stiffness value.

\end{appendices}


    \bibliographystyle{IEEEtran}
    \bibliography{bib/IEEEabrv,bib/mine,bib/iros2019_rlstrip}
\end{document}